# Simple Cortex: A Model of Cells in the Sensory Nervous System

**David Di Giorgio**

neuroowl.com/ddigiorg

## Abstract

Neuroscience research has produced many theories and computational neural models of sensory nervous systems. Notwithstanding many different perspectives towards developing intelligent machines, artificial intelligence has ultimately been influenced by neuroscience. Therefore, this paper provides an introduction to biologically inspired machine intelligence by exploring the basic principles of sensation and perception as well as the structure and behavior of biological sensory nervous systems like the neocortex. Concepts like spike timing, synaptic plasticity, inhibition, neural structure, and neural behavior are applied to a new model, Simple Cortex (SC). A software implementation of SC has been built and demonstrates fast observation, learning, and prediction of spatio-temporal sensory-motor patterns and sequences. Finally, this paper suggests future areas of improvement and growth for Simple Cortex and other related machine intelligence models.

## 1  Introduction

Artificial intelligence research has been built, to varying degrees, on a foundational understanding of biological sensory nervous systems. Popular machine learning techniques like artificial neural networks (ANNs) and deep learning were originally inspired using a dramatically simplified point neuron model **[1]**. Although architecturally inspired by biology, these ANNs use brute force function optimization of a cost function and backpropagation of errors to modify neural weights and successfully learn data. Some hypothesize backpropagation has a basis in neuroscience **[2]** while others explore a similarities between biological spike-time-dependant-plasticity and unsupervised learning mechanism **[3]**. Due to recent hardware advances and refined algorithms, math-heavy optimization based machine learning has become very successful. Extensive development of the mathematics of efficient optimization has yielded, for instance, interesting AI projects like Deep Mind's AlphaGo **[4]** (beating a world champion in the game of Go), and various self driving car solutions (along with many other world changes examples). While these advancements and projects have thus far facilitated a booming machine learning industry, leading experts like Geoffrey Hinton have begun to doubt math based learning methods like backpropagation will continue to significantly advance the AI field **[5]**.

Others researchers have a more neuroscientific focused perspective towards understanding and applying intelligence. A machine intelligence company, Numenta, heavily applies neuroscience research towards building a theoretical framework of the neocortex, Hierarchical Temporal Memory (HTM) **[6] [7] [8]**. Numenta has built a software suite, NuPIC **[9]**, that yields strong results in high-order sequence learning **[10]** compared to standard machine learning long short-term memory (LSTM) systems. HTM's development has yielded capable spatial and temporal anomaly detection in noisy domains **[11]** as well as a growing community of neuroscience and AI researchers. Many of the ideas and concepts regarding the neocortex consolidated by Numenta are discussed in section 2 The Neocortex.

Other researchers take a more unique approach towards understanding the computational processes of the brain. Ogma Intelligent Systems Corp introduces a model, Feynman Machine, that links neuroscience and mathematics by developing a learning architecture based on neuroscience and the applied mathematics of interacting dynamical systems **[12]**. Ogma has developed a library, OgmaNeo **[13]**, which has successfully been applied towards a simple self driving car solution **[14]** on a raspberry pi.

Not only are companies developing software models, others are extending neuroscience knowledge towards hardware solutions. Intel's self-learning neuromorphic chip, Loihi, is inspired by neural spikes, sparsity, and synaptic plasticity that can be modulated based on timing **[15]**. Intel plans to start manufacturing these chips and share it with university and research institutions.

Finally, there are researchers who are not associated with academia or industry that provide interesting insights into neuroscience and intelligence research. Louis Savian's Rebel Cortex **[16]** and his blog **[17]** state that the brain senses, learns, and perceives the world based on precise timing of neural spikes, a concept discussed in sections 1.1 Sensation and 1.2 Perception and used by Simple Cortex.

At their core, all of the researchers and their ideas discussed in this introduction have been influenced to varying degrees by the principles of intelligence found in biology. As our understanding of sensory nervous systems like the brain become more sophisticated, we can apply that knowledge towards building more general and capable intelligent machines.

## 1.1  Sensation

An intelligent entity exists, observes, and interacts in an environment, a place where concepts have causal relationship to each other. Humans call our environment the "Universe", where energy and matter exist and behave through natural laws applied to concepts like "space", "time", "gravity", "electromagnetism", "quantum physics", etc. These laws give rise to, cause, or explain relatively stable spatial structures and recurring temporal events within the Universe. Due to this causal natural order of the environment, biological nervous systems are able to witness and react to recurring knowledge (patterns and sequences) in the environment.  Therefore intelligent entities use large networks of sensors to sense, perceive, predict, and learn patterns and sequences from the environment as well as influence the environment through patterns and sequences of motor actions.  Thus a fundamental understanding of sensation is a good starting point for understanding the basic principles of intelligence.

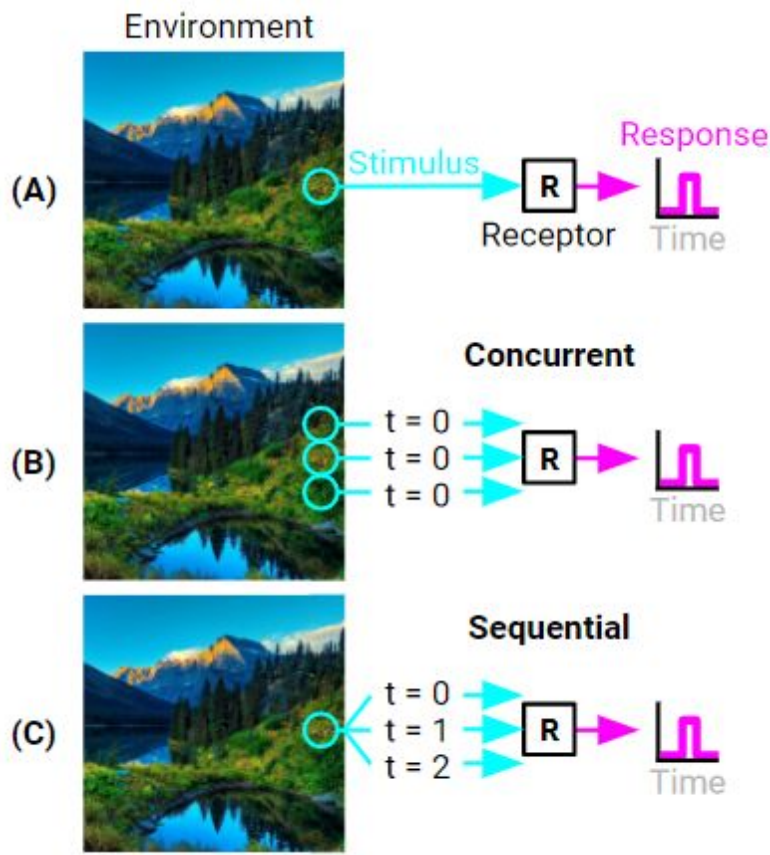

**Figure 1.1** - **(A)** When a receptor receives a specific stimulus (i.e. green light) it outputs a response signal (i.e. a spike of electrical potential).  **(B)** Concurrent stimuli are changes that occur within the same time range causing a receptor to respond.  **(C)** Sequential stimuli are changes that occur during different time ranges, usually proceeding one after another, causing a receptor to respond.

Sensation is simply the detection of stimulation in an environment.  A stimulus (plural stimuli) is a detectable phenomenon.  Examples include light frequency (color), sound frequency (pitch), chemicals (tastes and smells), and spikes of voltage (electrical action potentials).  A receptor is a structure that receives streams of stimuli and outputs a signal when it detects specific stimuli.  Examples include photoreceptors (vision), mechanoreceptors (hearing and touch), chemoreceptors (taste and smell), and synapses (electrical action potentials). A receptor's output signal is a discrete temporal marker **[16]** indicating the occurrence of specific phenomenon in a stream within a range of time. For example, take a receptor that receives one stream of stimuli and only outputs a signal when it detects blue light. Initially the receptor receives red light for a time, but eventually it receives a burst of blue light. This color change causes the receptor to output a signal, announcing the blue stimulus was recognized. Thus the most basic unit of sensory systems are sensory receptors that output a signal when they detect specific stimuli.

Receptors that receive multiple sensory streams operate purely on temporal relationships. They do so in two ways: concurrently or sequentially **[16]**. Concurrent stimuli are multiple detectable phenomena in an environment that occur within a single time range. For example, a computer screen is a 2D array of pixels. The light emitted from this array hits your retina within a very narrow range in time, allowing the brain to recognize spatial objects. Sequential stimuli are multiple detectable phenomena in an environment that occur during different time ranges, usually proceeding one after another. Continuing the example of the computer screen, the changes of a pixel from red to green to blue are sequential stimuli.

## 1.2  Perception

Perception is the basis of prediction, expectation, sophisticated object recognition (occlusion, invariance, and discrimination), and attention. Perception functions like sensation (response to detected phenomenon), however it utilizes stimuli from the memories of the sensory nervous system to make sense of the observed stimuli. For example, perception ("top-down" processing) is understanding you are seeing a chair, but sensation ("bottom-up" processing) is simply your eyes detecting light waves. Other intuitive examples of perception are found in optical illusions, which demonstrate the brain uses memories to make sense of the environment, as shown in Figure 1.2. Thus a receptor "perceives" rather than "senses" when it receives stimuli from an internal sensory receptor that doesn't directly observe the external environment.

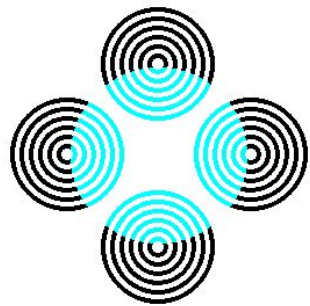

**Figure 1.2** - Neon color spreading optical illusion. The light cyan circle does not exist, however the brain perceives it is there based on the pattern created by the cyan arc stimuli.

## 2 The Neocortex

The mammalian neocortex is considered the highest seat of intelligence and is primarily responsible for producing humanity's great technological, philosophical, and cultural works. The human neocortex is a sheet of neural tissue occupying about 75% of the brain's volume, is around 2-4 mm thick, and is about 2,500 square cm in area (roughly the size of a dinner napkin) **[6]**. It is wrapped around older brain structures and the wrinkles, its defining physical characteristic , help increase the neocortical surface area within the skull. Just one square mm of this neural sheet has about 100 thousand neurons and 1 billion synapses. The neocortex is distributed in six layers, containing additional sublayers, based on differences in the relative density of neurons, axons, and synapses as well as interconnections to the rest of the brain **[6]**. Despite structural differences found in neocortical regions (mostly different neuron types and more or less layers/sublayers) the neocortex is remarkably homogeneous in the behavioral principles of cells and cortical areas **[6]**. By understanding how sensation and perception work within these areas, we gain valuable insight into building more capable models and explaining how existing models work.

### 2.1 Neocortical Structures

Sensory nervous systems are complex networks of interlinked sensors called "cells", which are collections of receptors forming a single sophisticated sensor that detects stimuli. The cells thought to be primarily responsible for intelligent behavior in the neocortex are called "neurons". A neuron has a cell body called the "soma", one or more branching structures called "dendrites" containing receptors called "synapses", and a tubular structure called the "axon" for outputting action potentials. A diagram of these structures are found in Figure 2.1. The most common type of neuron in the neocortex is the pyramidal neuron **[7]** which has thousands of dendrites forming synaptic connections to axons of other cells in various areas internal or external to the cortical hierarchy **[7]**. Synapses are receptors that detect stimuli, covering dendrites much like the leaves of a tree. These structures are thought to be the most fundamental memory storage and learning unit of the neocortex **[6]**. Thus dynamically interacting stimulus-receptor networks like the sensory nervous system are built from these axon-synapse connections.

Dendrites are the physical inputs of a neuron (the word "dendrite" comes from the Greek word "dendros" meaning "tree", describing its branch-like structure). As found in pyramidal neurons and other types of neurons, dendrites branch towards many areas of the sensory nervous system both within and outside of the neocortex  These different areas will contain stimuli representing different contextual information about the environment or an observed phenomenon. Take V1 for example, the primary visual processing area of the neocortex found near the back of the skull. V1 neurons have dendrites forming connections within the thalamus, a structure that receives information from environment-observing sensory systems like the retina and cochlea **[7]**. V1 neurons also have dendrites forming connections to higher brain regions like V2 and V3, which contain more sophisticated perceptual information. This is true within other areas of the neocortex. Generally, pyramidal neurons found in layer 3b of many brain areas have three types of dendrites: proximal, basal (or distal), and apical **[7]**. Proximal dendrites respond to bottom-up stimuli from the thalamus. Basal dendrites respond to lateral stimuli from neurons in the same cortical area **[7]**, which is thought to be the basis of cortical sequence learning. Apical dendrites, responding to top-down stimuli from higher cortical areas **[7]**, provide more sophisticated context, expectation, perception, and prediction about an observed phenomenon. Thus by branching dendrites to various areas of the sensory nervous system, a neuron can observe and recognize extremely complex patterns of stimuli, forming a sophisticated understanding of the environment.

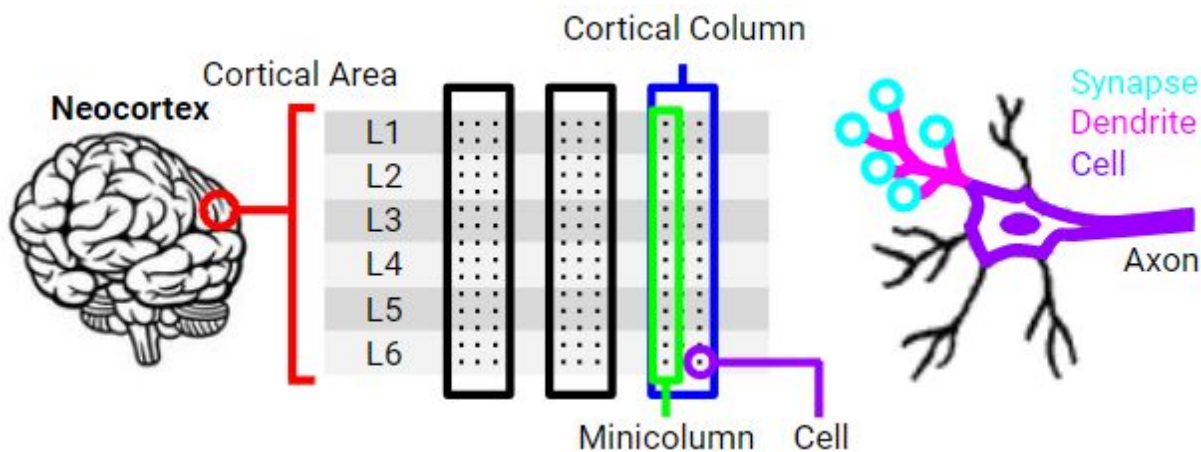

**Figure 2.1** - The mammalian neocortex consists of a hierarchy of interacting cell clusters called "cortical areas" or "brain regions". These areas are subdivided by 6 layers and consist of structures called "cortical columns" (or "macrocolumns", "minicolumns", and "cells". The cells that make up these columns have input structures called "dendrites" and "synapses" and an output structure called the "axon".

## 2.2  Neural Behavior

A neuron operates on the principles described in section 1.1 Sensation. A synapse is a receptor that receives streams of excitatory or inhibitory neurotransmitter stimuli from other neural axons. Dendrites are a collection of synapses whose responses within a time range sum up to a single input dendritic response which affects the parent neuron's state. When a single dendrite receives enough excitatory neurotransmitters through its synapses it depolarizes the parent neuron's soma, putting the neuron in an "predictive" or "expective" state. As the neuron continues to receive more excitatory dendritic responses, the soma depolarizes to a point where it fires action potentials (spikes or pulses of electrical potential) along the neuron's axon. These action potentials are the neuron's output response signal, temporal markers indicating the occurrence of enough specific stimuli within a time range. Conversely when a neuron receives enough inhibitory synaptic neurotransmitter input, it is repolarized and put in an inactive state (no spikes or spiking very slowly). Neural inhibition is an important principle of sensory nervous systems. It has been observed among all areas within the neocortex the activity of neurons is "sparse" **[6]** (i.e., from a population of neurons only a very small percentage are active in a given moment while the rest are inactive). Thus inhibition is how sensory nervous systems process and learn only the most important information from a complex dynamic environment.

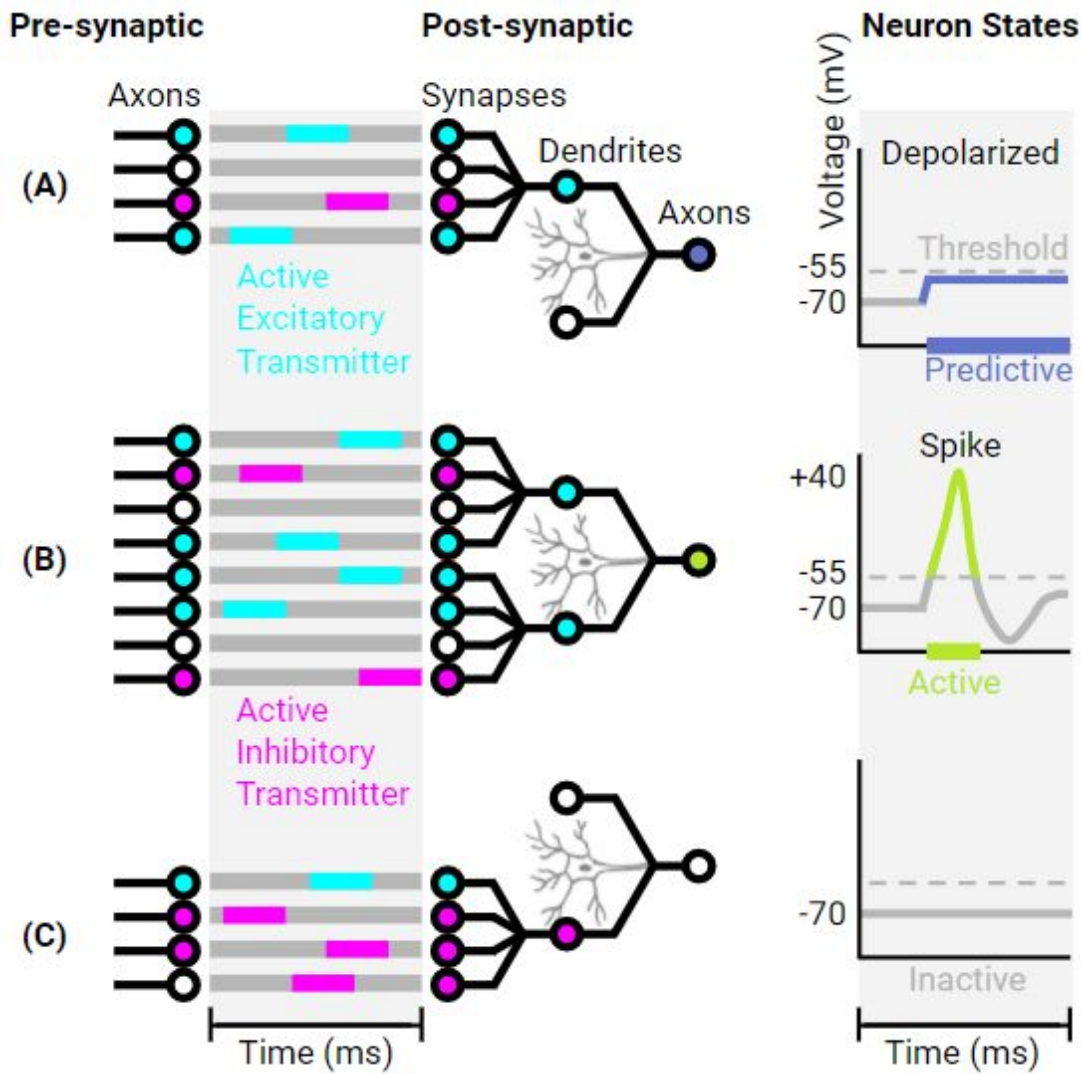

**Figure 2.3** - Example of neural sensation. Excitatory (cyan) and Inhibitory (magenta) neurotransmitters are released from pre-synaptic axons within a time range. **(A)** A single dendrite has received enough excitatory input to depolarize the neuron, putting it in a "predictive" state. **(B)** Multiple dendrites have received enough excitatory input to activate the neuron which outputs action potentials. **(C)** Enough dendrites have received enough inhibitory input to suppress the neuron's activation.

## 2.3 Synaptic Plasticity

Synapses form memories from observed knowledge simply by growing towards active stimuli and shrinking away from inactive stimuli **[7]**.  This effect is called Hebbian Learning, a dendritic and synaptic behavior much like plants that grow towards light for sustenance and shrink away from darkness.  As stimuli evolves over time, synapses connected to constantly recurring active stimuli strengthen and become more permanent while synapses connected to rarely recurring active stimuli weaken and become more easily forgotten.  It has been observed that synapses form and reform within a span of milliseconds (short term memory) **[6]** while some synaptic connections last an entire lifetime (long term memory).  It is from this simple behavior that biological sensory nervous systems adapt their memories to a dynamically changing complex environment.

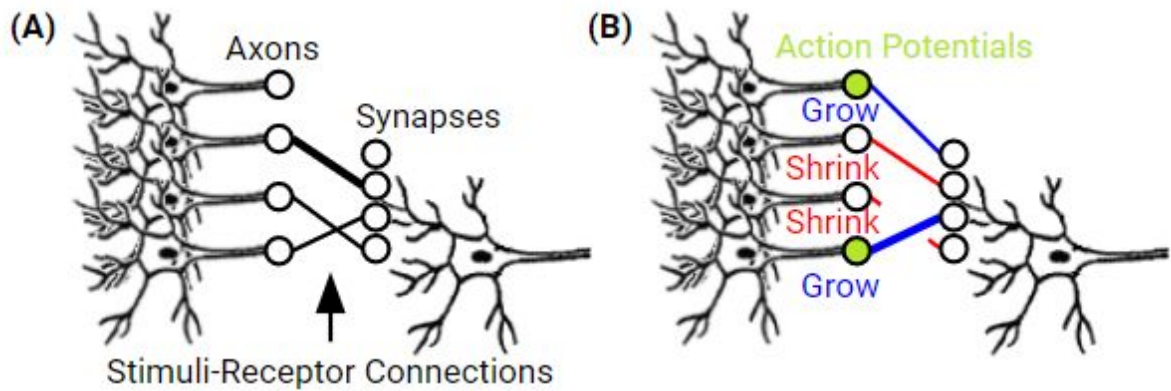

**Figure 2.4** - Example of synaptic learning between neurons.  **(A)** The synapses of the post-synaptic neuron have previously formed connections with the axons of pre-synaptic neurons.  The thickness of the connection, indicated by the line between axon and synapse, represents the connective permanence (or strength) of the synapse.  **(B)** When the pre-synaptic neurons activate and output action potentials, the post-synaptic neuron's synapses grow towards active axons and shrink away from inactive axons.

## 3   Simple Cortex Model

The Simple Cortex model of neurons is based on the sensory-receptor concepts discussed in Section 1.1 Sensation and neocortical structures and behavior discussed in Section 2 The Neocortex.  Concepts like spike timing, synaptic plasticity, inhibition, and neural structures are applied to SC.  The following terms are used in the architecture of this model in Section 3.7 Architecture.

## 3.1 Stimuli

A SC stimuli is simply the input(s) and/or output of SC neurons. Examples include data from sensors, motor/actuators, and action potentials (active neurons) or depolarized somas (predicted neurons). Stimuli data is stored and processed using one data vector:

- **sStates**: "Stimuli States" is a binary vector where each value represents whether a stimulus is "active" or "inactive".

## 3.2 Synapse

A SC synapse is a receptor that responds to stimulus and has memory. A synapse behaves using Hebbian-like synaptic learning rules (inspired by HTM) found in Section 4 Algorithms of this paper. Because neuroscience suggests the synapse is the most fundamental memory storage of the neocortex, the synaptic data is a model of observed stimuli over time. Synapse data is stored and processed using two vectors:

- **sAddrs**: "Synapse Addresses" is vector where each value represents where a synapse is connected to a stimulus location.

- **sPerms**: "Synapse Permanences" is a vector where each value represents how strongly a synapse is connected, from 0 to 99.

## 3.3 Dendrite

A SC dendrite is a collection of synapses with a threshold value. When enough synapses are connected to active stimuli the dendrite activates, representing the occurrence of a recognized pattern. The threshold value is a user-defined percentage of the total number of synapses on the dendrite. Threshold percentages below 100% allow for pattern recognition in noisy environments because only a subset of learned stimuli needs to be present to activate the dendrite. Dendrite data is stored and processed using one vector and one value:

- **dOverlap**: "Dendrite Overlaps" is a vector where each value represents how many synapses on a dendrite are connected to active stimuli during a time step.

- **dThresh**: "Dendrite Threshold" is an value that represents the minimum dendrite overlap value required to activate the dendrite.

## 3.4 Forest

A SC forest is a collection of one dendrite from all neurons in a SC area. It is used to organize data buffers for more efficient computational parallelization and storage of synaptic memories. SC forests represent dendrites in a cortical area that respond to stimuli that share the same receptive field. For example, a SC area may have 100 neurons with 5 dendrites each. In this example, all first dendrites of the 100 neurons form a forest where these dendrites only respond to a window of pixels in the top left of an input space. The synapse addresses and permanences of these dendrites would be stored in one sAddrs buffer and one sPerms vector and these vectors would respond to and learn from one stimuli buffer. Continuing the example, all second dendrites of the 100 neurons would form another forest, and so on.

## 3.5 Neuron

A SC neuron is a collection of dendrites with a threshold and models the output of action potentials (neuron activations) and soma depolarization (neuron predictions). A SC neuron learns and responds to dendrite activation patterns that happen coincidentally or sequentially. An example of neuron activation due to coincident stimuli is shown in the Encode and Learn steps in Figure 4.1 found in section 4 Algorithms. Once a coincidence or sequence has been learned, a neuron can use the occurrence of one or more learned dendrite activations to enter a predictive state. A neuron in a predictive state implies the occurrence a dendritic pattern, even if that dendrite was not activated through observed stimuli. An example of neuron prediction is also shown in the Predict and Decode steps in Figure 4.1. Neuron data is stored and processed using three vectors and one (or more) value(s):

- **nOverlaps**: "Neuron Overlaps" is a vector where each value represents how many dendrites on a neuron are active during a time step.

- **nStates**: "Neuron States" is a binary vector where each value represents whether a neuron is "active" or "inactive".

- **nBoosts**: "Neuron Boosts" is a vector where each value represents how often a neuron is inactive. In the SC algorithms, neurons activated less frequently are more likely to become active when learning new patterns.

- **nThresh**: "Neuron Threshold" is a value that represents the minimum neuron overlap value required to activate the neuron. Additionally, there may be another threshold value that represents the minimum neuron overlap value required to predict the neuron.

## 3.6   Area

A SC area is a collection of neurons governed by activation, inhibition, and boosting rules found in section 4 Algorithms.  Confident neurons, those that have many active dendrites, inhibit the activation of uncertain neurons, those that have few to no active dendrites.  If there are no confident neurons, a completely new pattern has been observed.  Therefore the neuron with the largest boost value, representing the neuron activated the least frequently, is activated so that it may apply the synaptic learning rules to memorize the new pattern.  Eventually every synapse in a SC area will have memories.  In this situation when a new pattern is observed, a SC area simply activates the neuron with the largest boost value and applies the synaptic learning rules.  Thus in SC only the most commonly recurring observed coincident or sequential patterns will be remembered while rare and irrelevant knowledge is forgotten.

## 3.7   Simple Cortex Architecture

The data vectors explained in this section use similar nomenclature found in neuroscience.  Figure 3.1 provides a visual example of a SC sequence learning model via neural output feedback.

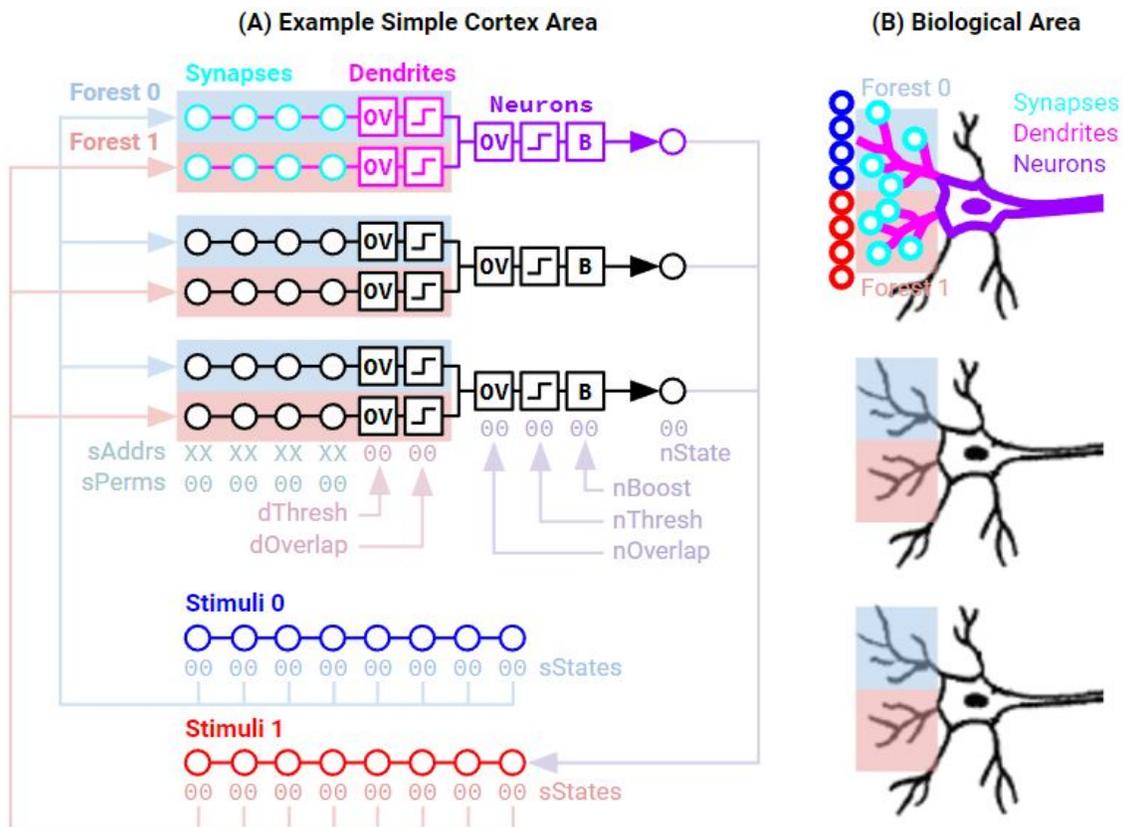

**Figure 3.1** - **(A)** A visual example of SC area containing a flow diagram of stimuli, synapses, dendrites, forests, and neurons. Synapses contained in the blue and red colored forests respond to stimulus in the their respective stimuli vector. If the dendritic overlap value exceeds the dendritic threshold (enough synapses detect their stimulus), the dendrite is considered "active". The same overlap process is done for neuron selection, but with inhibition and boosting rules explained in Section 3.5 Neuron. Note, neuron states are fed back into stimuli 1. This feedback is not required in every SC implementation, but its use is crucial to learning sequences of neuron activations. Additionally, SC is not limited to the amount of structures shown in the example and therefore may have as many stimuli, synapses per dendrite, dendrites per neuron, and neurons as required. **(B)** For comparison, a visual example of biological neurons has been provided.

# 4 Simple Cortex Algorithms

The Simple Cortex algorithms are based on the sensory-receptor concepts discussed in Section 1 Introduction and neocortical structures and behavior discussed in Section 2 The Neocortex. Concepts like spike timing, synaptic plasticity, inhibition, and neural structures are applied to SC. The following terms are used in Section 3.7 Behavior.

## 4.1 Set Stimuli

The SC Set Stimuli algorithm is equivalent to HTM Theory's Encode algorithms. Both simply convert scalar arrays or numbers like pixel values or audio frequency to binary vectors representing "active" and "inactive" states.

## 4.2 Encode

The SC Encode algorithm is fundamentally a pattern recognizer. It outputs a binary vector representing "active"/"inactive" neuron states by observing stimuli vectors and comparing them with the architecture's synaptic memories stored in forests. Encode consists of four steps:

1. For every forest, overlap synapses with stimuli. See "overlapSynapses" in Supporting Information section for pseudocode details.

2. Activate neurons using the previously calculated dendrite overlap values. Increment all boost values. See "activateNeurons" in Supporting Information section for pseudocode details.

3. If there are no confident neurons, meaning no inhibition occurs, activate neurons with highest boost value.

4. Zero active neuron(s) boost values.

## 4.3 Learn

The SC Learn algorithm occurs after the Encode step activates neurons. It uses a modified Hebbian Learning style rules inspired by HTM Theory by updating synapse address and permanence values from the observed stimuli in the Encode step. See "learnSynapses" in Supporting Information section for pseudocode details. Learn consists of one involved step:

1. For every active neuron, perform the following learning rules:

- **Grow Synapses**: Increment the synapse permanence if the synapse address of its connected stimulus is active. Synapses that "grow" are connected to a stimulus that occurred when expected. Larger permanence values represent a more lasting connection with the stimulus stored in the synapse address.

- **Shrink Synapses**: Decrement the synapse permanence if the the synapse address of its connected stimulus is inactive. Synapses that "shrink" are connected to a stimulus that did not occur when expected. Smaller permanence values represent a synapse connection highly susceptible to moving to a new stimulus.

- **Move Synapses**: If a synapse permanence is 0 (representing an unconnected synapse), set the stimulus address to an unused stimulus and reset the synapse permanence to 1. Therefore, synapses "move" towards recurring active stimulus and away from inactive stimulus.

## 4.4 Predict

The SC Predict algorithm functions like the Encode algorithm, but without the neuron inhibition and boosting rules. It outputs a binary vector representing "predict"/"inactive" neuron states by observing one or more stimuli vectors and comparing them with the architecture's synaptic memories stored in forests. Predict consists of two steps:

1. For each specified forest, overlap synapses with stimuli. See "overlapSynapses" in Supporting Information section for pseudocode details.

2. Predict neurons using the previously calculated dendrite overlap values. See "predictNeurons" in Supporting Information section for pseudocode details.

## 4.5 Decode

The SC Decode algorithm converts neuron activations or predictions back to stimuli patterns using synapse memories. Decode consists of one step:

1. Activate stimuli based on the synapse addresses of active neurons. See "decodeNeurons" in Supporting Information section for pseudocode details.

## 4.6 Simple Cortex Behavior

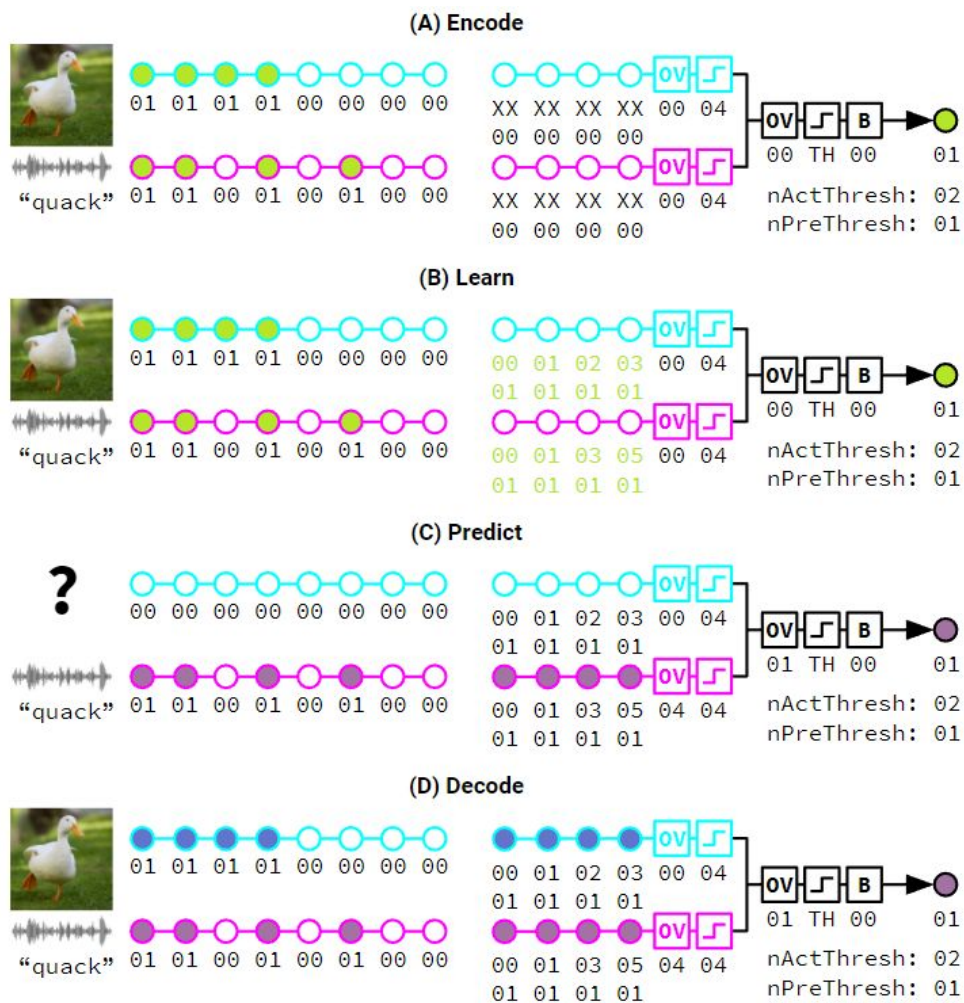

**Figure 4.1** - Example of SC algorithms using an architecture with 2 stimuli, 2 dendrites per neuron, and 1 neuron per area. **(A)** Stimuli 0 is a binary vector representing a picture of a duck and stimuli 1 is a binary vector representing a duck's "quack" sound. Encode algorithm is run on a neuron with no memories (unconnected address and zero permanence). Neurons are activated based on the rules discussed below.. **(B)** Learn algorithm modifies the synapses of active neurons based on the rules discussed below **(C)** Predict algorithm selects neurons based on learned knowledge. **(D)** Decode algorithm activates stimuli from the synapse addresses of predicted neurons.

# 5   Results

The Simple Cortex architecture and algorithms detailed in this paper have been implemented in C++ using OpenCL 1.2 for use on both CPU and CPU/GPU platforms. The implementation runs on Windows and Linux machines and is available at https://github.com/ddigiorg/simple-cortex.

A demo has been created to demonstrate Simple Cortex's ability to recognize, learn, and predict sequences of observed patterns. It contains a 100x100 pixel environment with a circular ball made of 49 pixels. The ball is governed by simple 2D physics rules: acceleration towards the ground due to gravity and energy loss due to bouncing off the ground and side walls. The ball may have random (X, Y) initial start positions across the screen and random (X, Y) initial velocities within a specified range of values.

A SC architecture has been created in the ball demo containing 1.5 million neurons with 2 dendrites each. The first dendrite of every neuron has 50 synapses each, is contained in Forest 0, and observes and learns from the input scene state stimuli vector. The second dendrite of every neuron has 1 synapse each, is contained in Forest 1, learns from the previous active neurons stimuli vector, and predicts the next time step's active neurons from the current active neurons stimuli vector. For every time step after Encode and Learn user may enable Forecast which loops Predict and Decode up to a specified amount. The length of the Forecast loop determines how far into the future SC predicts the ball's potential trajectory. See "Ball Demo" in Supporting Information section for pseudocode details.

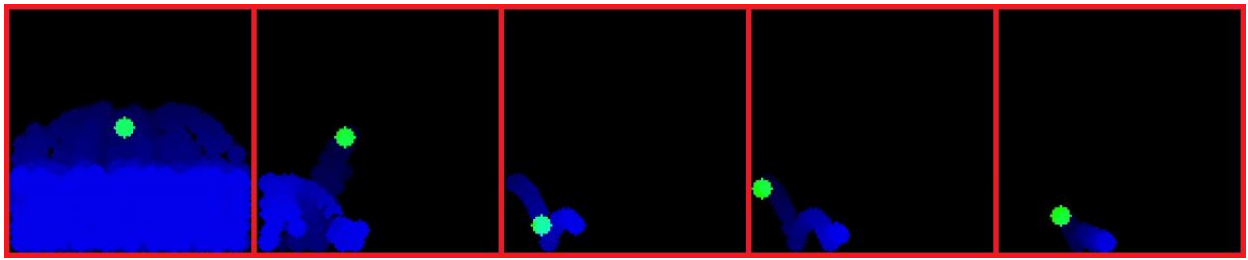

**Figure 5.1** - Sample frames of Ball Demo showing Simple Cortex algorithms in action.  Green pixels represent input stimuli and blue pixels represent decoded predictions of ball trajectories up to 20 time steps into the future.  In the first example frame SC has no context where the ball is traveling so it predicts all trajectories it has observed based on past experience.  The 2nd example frame shows that SC has more temporal context about the ball's travel direction, so the predicted trajectories are refined.  By the 3rd example frame SC is predicting only 1 trajectory, which turns out to be accurate based on the 3rd and 4th example frames.

Ball Demo is benchmarked for speed using these parameters:

- **Compute Device**: Nvidia GTX 1070 GPU
- **Samples**: Mean and Standard Deviation values used 100,000 time step samples
- **Neurons**: 1.5 million with 2 dendrites each
- **Forest 0**: 50 synapses per dendrite, 25% activation threshold (13 active stimuli required to activate dendrite)
- **Forest 1**: 1 synapse per dendrite
- **Total Synapses**: 76.5 million
- **Forecasting**: 20 predicts/decodes per time step

| Executed Algorithms | Mean (ms) | Std. Dev. (ms) |
|---|---|---|
| **Encode** | 6.72 | 0.00[1] |
| **Encode, Learn** | 7.44 | 4.24[2] |
| **Encode, Learn, x20 Predict, x20 Decode** | 21.76 | 4.47[2] |

**Table 5.1**: Computation speeds (mean and standard deviation) per frame. **(1)** Encode algorithm had S.D. less than 1/100th of a millisecond and therefore written as 0.00 ms. **(2)** Learn algorithm causes high standard deviation due to unoptimized "move synapse" algorithm.

Based on the numbers in Table 1, Simple Cortex can encode and learn ~10 billion synapses/second according to compute speed benchmark (76.5 million synapses in 7.44 ms).  The speed of recognizing patterns each frame is very consistent, however updating the synapses may cause slowdowns, probably due to the "move synapses"portion of the Learn algorithm.

Simple Cortex: A Model of Cells in the Sensory Nervous System

## 6   Conclusions & Future Work

This paper has discussed the principles of sensation and neocortical intelligence. Additionally, it has presented a new straightforward neural model based on the architecture and behavior of the neocortex. A Simple Cortex implementation has been built to observe, learn and predict spatio-temporal sensory-motor patterns and sequences. Simple Cortex is:

- **Simple**: No sophisticated knowledge of mathematics required

- **Unsupervised**: Stores knowledge through observation with no labeled data necessary

- **On-line**: Observes and learns continuously like biological intelligence

- **Dynamic**: memories adapt to new observations, prioritizing commonly recurring knowledge

- **Predictive**: SC is able to predict neuron states many time steps into the future. It can also translate these neuron states back to observable data.

- **Fast**: GPU parallel processing encodes and learns billions of synapses per second.

However, Simple Cortex is still in its infancy and has a lot of room for growth. Here are a few ideas, among many, for improvement and future investigation:

- It would be extremely beneficial to benchmark SC to NuPIC, OgmaNeo, and LSTM on sequence learning tasks. Additionally, SC should be compared to traditional Deep Learning techniques on well known benchmarks like MNIST for handwritten digit recognition.

- Modify SC to process scalar stimuli data vectors like images with color pixel values. In its current implementation SC only observes, learns, and predicts binary input stimuli. This limits the implementation only to binary data sets.

- Create demos applied to stimuli like movement-receptive retinal ganglion cell models for advanced biologically-based object tracking.

- SC algorithms and it's code implementation would benefit from optimization improvements, especially in the "move synapses" portion of the Learn algorithm, the biggest speed bottleneck.

Simple Cortex: A Model of Cells in the Sensory Nervous System## Supporting Information

### Simple Cortex Basic Functions

**overlapSynapses** - This function modifies neuron overlap buffer. For each neuron, every dendrite's overlap is calculated by determining if enough synapses are connected to active stimuli. If the dendrite overlap value beats the dendrite threshold value, then increment the neuron overlap value.

```
00  for n in neurons
01    dendriteOverlap = 0
02
03    for s in neuronSynapses
04      if synapsePermanances[s] > 0 and stimuliStates[synapseAddresses[s]] > 0
05        dendriteOverlap++
06
07    if dendriteOverlap >= dendriteThreshold
08      neuronOverlaps[n]++
```

**activateNeurons** - This function modifies neuron states buffer, neuron boosts buffer and inhibition flag value. For each neuron, increment the boost value. If the neuron overlap value beats the neuron activation threshold value (and therefore is considered a confident neuron), then zero the neuron boost value, set the neuron state to "active", and set inhibition flag true.

```
00  for n in neurons
01    if neuronBoosts[n] < synapseAddressMaximum
02      neuronBoosts[n]++
03
04    if neuronOverlaps[n] >= neuronThresholds
05      neuronBoosts[n] = 0
06      neuronStates[n] = 1
07      inhibitionFlag = true
```

David Di Giorgio                                                                                    18/22

**learnSynapses** - This function modifies synapse address buffer and synapse permanence buffer by using the synaptic learning rules explained in section 4.3 Learn. This function only applies to active neurons. Note: the move portion of this function is unoptimized.

```
00  for n in neurons
01    if neuronStates[n] > 0
02
03      // Grow and or Shrink
04      for s in neuronSynapses
05        if synapsePermanences[s] > 0
06          if synapseStates[synapseAddresses[s]] > 0
07            if synapsePermenences[s] < synapsePermenenceMaximum
08              synapsePermenences[s]++
09          else
10            synapsePermenences[s]--
11
12      // Move
13      j = 0
14
15      for s in neuronSynapses
16        if synapsePermenences[s] == 0
17          for i starting at j in numStimuli
18            if stimuliStates[i] > 0
19              usedStimulusFlag = true
20
21              for s2 in neuronSynapses
22                if synapseaddresses[s2] == i
23                  usedStimulusFlag = false
24                  break
25
26              if usedStimulusFlag == true
27                synapseAddresses[s] = i
28                synapsePermenences[s] = 1
29                j = i + 1
30                break
```

**predictNeurons** - This function modifies the neuron states buffer. For each neuron, if the neuron overlaps value beats the neuron predicted threshold value, set the neuron states to "predicted".

```
00  for n in neurons
01    if neuronOverlaps[n] >= neuronThreshold
03      neuronStates[n] = 1
```

**decodeNeurons** - This function modifies a specified stimuli states vector by setting the stimulus address "active" for every synapse address in a forest as long as its neuron is predicted.

```
00  for n in neurons
01    if neuronStates[n] > 0
01      for s in neuronSynapses
02        if synapsePermanances[s] > 0
03          stimuliStates[synapseAddresses[s]] = 1
```

## Ball Demo

Use the example SC area in Figure 3.1 as a reference. Stimuli 0 is a binary data vector representing "active" or "inactive" ball scene pixels (line 3). It is encoded and learned in forest 0 (lines 5 and 6). Stimuli 1 is a binary vector representing the previous "active" or "inactive" neuron states (line 8). It is encoded and learned in forest 1, (lines 5 and 6). The code then loops for as many future time steps as specified (line 10). Stimuli 2 stores the current neuron states (line 11), which is used with forest 1 to predict the next time step's neuron states (line 13). Stimuli 3 stores the next time steps active stimuli, which is decoded from the synapses of predicted neuron states of forest 0 (line 14). These predictions are stored and remembered (lines 16 - 18) for each forecast loop.

```
00  loop
01    ballScene.update()
02
03    stimuli[0].set(ballScene.data)
04
05    area.encode({stimuli[0], stimuli[1]}, {forest[0], forest[1]})
06    area.learn({stimuli[0], stimuli[1]}, {forest[0], forest[1]})
07
```

```
08      stimuli[1].set(area.nStates)
01
10      for f in numForecasts
11        stimuli[2].set(area.nStates)
12
13        area.predict({stimuli[2]}, {forest[1]})
14        area.decode({stimuli[3]}, {forest[0]})
15
16        for p in numPixels
17          if stimuli[3].sStates > 0
18            prediction[p] = 1
```

## Acknowledgements

I thank Jeff Hawkins[1] and Numenta for their work on neocortical intelligence and NuPIC. I thank Jacob Everist[3],Jake Bruce[3] and other collaborators for their comments and ideas. I thank Fergal Byrne[2] and Eric Laukien[2] for answering questions about Feynman Machine. I thank Louis Savien[4] for our discussions on neocortical intelligence. Finally I thank my friend and mentor, Frank Foote, for helping improve this paper.

[1]Numenta, [2]Ogma Intelligent Systems Corp., [3]HTM Forum (discourse.numenta.org), [4]Rebel Science

## References

[1]   McCulloch, W. S., and Pitts, W. (1943). A logical calculus of the ideas immanent in nervous activity. The bulletin of mathematical biophysics 5(4):115–133.

[2]   Marblestone, A. H. , Wayne G., and Kording K. P. (2016). Toward an Integration of Deep Learning and Neuroscience. Frontiers. doi: 10.3389/fncom.2016.00094

[3]   Bengio, Y., Lee, D-H., Bornschein, J., Mesnard, T., Lin, Z. (2016). Towards Biologically Plausible Deep Learning. arXiv:1502.04156v3

[4]   Silver, D., Huang, A., Maddison C. J., Guez, A., Sifre, L., Driessch G., et. al. (2016) Mastering the game of Go with deep neural networks and tree search. Nature 529, 484 - 489. doi: 10.1038/nature16961

Simple Cortex: A Model of Cells in the Sensory Nervous SystemSimple Cortex: A Model of Cells in the Sensory Nervous System

[5]   Geoffry Hinton's quote found online at:
       https://www.axios.com/ai-pioneer-advocates-starting-over-2485537027.html

[6]   Hawkins, J. et al. (2016). Biological and Machine Intelligence.  Release 0.4. Accessed at
       http://numenta.com/biological-and-machine-intelligence/

[7]   Hawkins, J., Ahmad, S. (2016).  Why Neurons Have Thousands of Synapses, a Theory of
       Sequence Memory in Neocortex.  Frontiers.  doi: 10.3389/fncir.2016.00023

[8]   Hawkins, J., Ahmad, S., Cui, Y. (2017).  Why Does the Neocortex Have Layers and Columns, A
       Theory of Learning the 3D Structure of the World.  bioRxiv. doi:
       https://doi.org/10.1101/162263

[9]   NuPIC implementation found online at: https://github.com/numenta/nupic

[10]  Cui, Y., Ahmad, S., Hawkins, J. (2016).  Continuous Online Sequence Learning with an
       Unsupervised Neural Network Model.  The MIT Press Journals, Volume: 28, Issue: 11, pp.
       2474-2504  doi: 10.1162/NECO_a_00893

[11]  Ahmad, S., Lavin, A., Purdy, S., Agha, Z. (2017).  Unsupervised real-time anomaly detection for
       streaming data.  ScienceDirect Volume 262, pp. 134-147. doi:
       10.1016/j.neucom.2017.04.070

[12]  Laukien, E. Crowder, R., Byrne, F. (2016).  Feynman Machine: The Universal Dynamical Systems
       Computer.  arXiv:1609.03971v1

[13]  OgmaNeo implementation found online at: https://github.com/ogmacorp/OgmaNeo

[14]  Self Driving Car demo found online at:
       https://ogma.ai/2017/06/self-driving-car-learns-online-and-on-board-on-raspberry-pi-3/

[15]  Intel Loihi article found online at:
       https://newsroom.intel.com/editorials/intels-new-self-learning-chip-promises-accelerate-artificial-intelligence/

[16]  Savien, L. (2011).  Rebel Cortex 1.0 A Visual Recognition Software Project.  Rebel Science.
       Available online at: http://www.rebelscience.org/download/rebelcortex.pdf
David Di Giorgio                                                                                                                      22/22